\documentclass[conference]{IEEEtran}
\IEEEoverridecommandlockouts
\usepackage{cite}
\usepackage{amsmath,amssymb,amsfonts}
\usepackage{algorithmic}
\usepackage{graphicx}
\usepackage{textcomp}
\usepackage{xcolor}
\usepackage{graphicx}
\usepackage{amsmath}
\usepackage{amssymb}
\usepackage{multicol}
\usepackage{multirow}
\usepackage{booktabs}
\usepackage{threeparttable}
\usepackage{array}
\usepackage{booktabs}
\usepackage{bbding}
 \usepackage{graphicx}
 \usepackage{amsmath}
 \usepackage{amsmath}
  \usepackage{amsfonts}
  \usepackage[mathscr]{euscript}
\def\BibTeX{{\rm B\kern-.05em{\sc i\kern-.025em b}\kern-.08em
    T\kern-.1667em\lower.7ex\hbox{E}\kern-.125emX}}
\begin{document}

\title{CLIPER: A Unified Vision-Language Framework for In-the-Wild Facial Expression Recognition
}

\author{Hanting Li, Hongjing Niu, Zhaoqing Zhu, Feng~Zhao$^{\dagger}$  \thanks{$^{\dagger}$ Corresponding to fzhao956@ustc.edu.cn} \thanks{H. Li, H. Niu, Z. Zhu , and F. Zhao are affiliated with University of Science and Technology of China, Hefei 230026, China.}}
\maketitle

\begin{abstract}
Facial expression recognition (FER) is an essential task for understanding human behaviors. As one of the most informative behaviors of humans, facial expressions are often compound and variable, which is manifested by the fact that different people may express the same expression in very different ways. However, most FER methods still use one-hot or soft labels as the supervision, which lack sufficient semantic descriptions of facial expressions and are less interpretable. Recently, contrastive vision-language pre-training (VLP) models (e.g., CLIP) use text as supervision and have injected new vitality into various computer vision tasks, benefiting from the rich semantics in text. Therefore, in this work, we propose \textbf{CLIPER}, a unified framework for both static and dynamic facial \textbf{E}xpression \textbf{R}ecognition based on \textbf{CLIP}. Besides, we introduce multiple expression text descriptors (METD) to learn fine-grained expression representations that make CLIPER more interpretable. We conduct extensive experiments on several popular FER benchmarks and achieve state-of-the-art performance, which demonstrates the effectiveness of CLIPER.
\end{abstract}

\begin{IEEEkeywords}
Facial expression recognition, vision-language model, prompt engineering
\end{IEEEkeywords}

\section{Introduction}
\label{sec:intro}

Facial expressions are one of the most common human behaviors that can reveal mental activities. Facial expression recognition (FER) is the premise of many important computer vision (CV) tasks, such as human-computer interaction \cite{liu2017hci}, driver safety monitoring \cite{wilhelm2019towards}, and healthcare aids \cite{bisogni2022impact}. Therefore, FER has become a popular task in recent years. 

\begin{figure}[t]

  \centering
  \includegraphics[width=6cm]{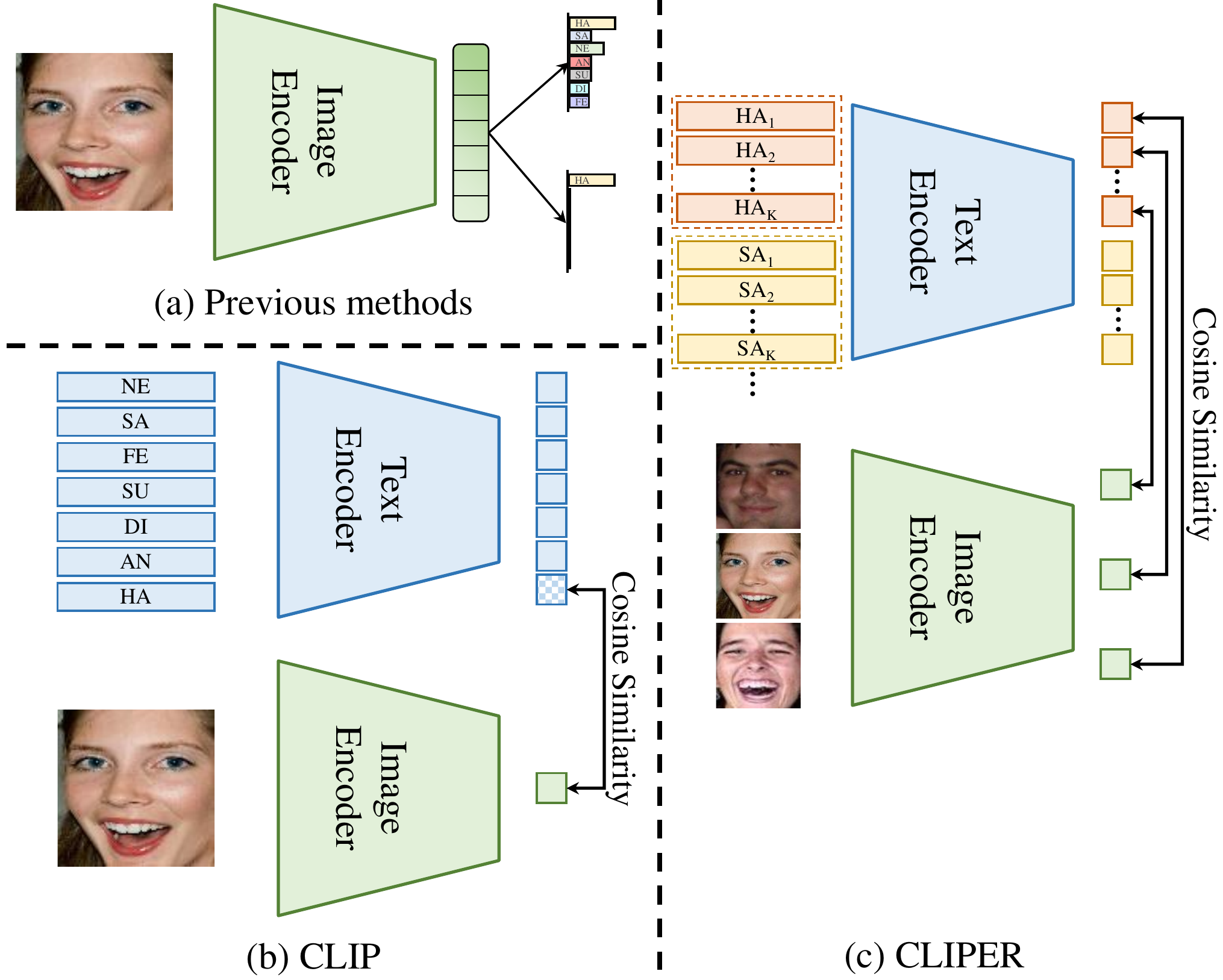}
  \caption{The structure of three facial expression recognition frameworks. (a) Most previous methods \cite{wang2020scn,zhao2021efficientface,ma2021vtff,xue2021transfer} use one-hot label or soft label to train the network, (b) standard CLIP \cite{radford2021clip} designs a text descriptor for each class, and use the cosine similarity between the image and text embeddings to make inference, and (c) our CLIPER generates $K$ text descriptors for each class, and use the mean cosine similarity between the image and a group of text embeddings to make the inference.}
 \label{Fig1}
\end{figure}

Among FER tasks under different scenarios, in-the-wild FER is more challenging than the laboratorial FER, due to its complex background, extreme poses, and random occlusions. Many FER methods have gradually improved the performance of static facial expression recognition (SFER) based on a single image \cite{wang2020scn,zhao2021efficientface,ma2021vtff,li2021mvt,xue2021transfer} and dynamic facial expression recognition (DFER) based on a video sequence \cite{zhao2021former,ma2022stt,wang2022dpcnet} . However, as shown in Figure~\ref{Fig1}(a), most current approaches are supervised by one-hot or soft labels that lack semantic information and interpretability, which undoubtedly hinders the further improvement of the FER methods.

Recently, large-scale contrastive vision-language pre-training (VLP) models (e.g., CLIP \cite{radford2021clip} and ALIGN \cite{jia2021align}) have introduced text to provide supervision with more semantic information for learning visual representations. The success of these VLP models inspires many researchers to transfer them to many downstream CV tasks and achieve promising results, such as video understanding \cite{luo2022clip4clip}, image synthesis \cite{patashnik2021styleclip}, and semantic segmentation \cite{xu2022groupvit}. Besides, studies have shown that these VLP models learn rich expression-related information \cite{goh2021clipvisual}. Therefore, using text to guide the image encoder to learn better facial expression representations is a very promising direction. 
Nevertheless, there are two main difficulties that need to be overcome when applying the VLP models to FER tasks. 1) Since the name of the classes in FER task (e.g., happy, sad, and angry) is abstract and not an object, the text prompt cannot be designed by using the standard prompt (i.e., ``a photo of a $\{\textbf{class}\}$''), as did in \cite{radford2021clip}. 2) Manually designing expression text prompts based on expert knowledge is cumbersome and unreliable, and it is difficult to design a uniform textual description for each expression (e.g., smiling and laughing can be two different forms of happiness). To overcome these two difficulties, we propose a vision-language facial \textbf{E}xpression \textbf{R}ecognition framework based on \textbf{CLIP} \cite{radford2021clip}, named \textbf{CLIPER}. It is worth noting that CLIPER can be applied to both SFER and DFER. Since the focus of this paper is not the aggregation of temporal information, we use the most straightforward and commonly used temporal average pooling to aggregate the features of each frame.

As shown in Figure~\ref{Fig1}(b), standard CLIP designs a handcrafted text descriptor for each class, which is not an optimal FER task for those expressions whose categories are compound and variable concepts. Specifically, the authors in~\cite{goh2021clipvisual} proved that most expression-related texts in CLIP, such as ``angry'' and ``surprise'', are compound concepts. For example, ``angry'' is highly related to ``evil'', ``serious'', and ``mental health disorder'', while ``surprise'' is strongly correlated with ``celebration''.

Therefore, we devise multiple expression text descriptors (METD), which can automatically learn a group of text descriptors for each expression that corresponds to different forms of the same expression, as shown in Figure~\ref{Fig1}(c). Our CLIPER not only achieves competitive results in terms of overall accuracy, but also obtains more fine-grained expression representations for each expression (e.g., ``a little happy'' and ``very happy''), which also indicates that our method is more interpretable. 

In summary, we make the following contributions:
\begin{itemize}
	\item We first introduce contrastive vision-language pre-training models to FER task and propose CLIPER, the first unified framework for both DFER and SFER.
	\item To avoid complicated manual design of text prompts and make CLIPER adapt to compound and variable facial expressions, we propose METD to automatically learn a group of fine-grained text descriptors for each expression, which brings interpretability to CLIPER.
	\item We conduct extensive experiments on five popular in-the-wild FER datasets. Our CLIPER achieves SOTA performance on these datasets and can serve as a strong baseline model for FER tasks. The visualization results demonstrate the interpretability of CLIPER. 
\end{itemize}

\section{Related Work}

\subsection{In-the-Wild Facial Expression Recognition}
As an essential task to understand human behavior and state, facial expression recognition has received increasing attentions in recent decades. In the beginning, handcrafted features, such as LBP \cite{shan2009lbp}, HOG \cite{dalal2005hog}, and SIFT \cite{ng2003sift}, are widely used to extract visual features from images or video sequences and achieve good performance on FER tasks .

Recently, deep-learning-based methods further improve the performance on both static and dynamic facial expression recognition tasks \cite{li2020fersurvey}. For SFER, researchers devise models based on convolutional neural networks (CNNs) \cite{wang2020scn,zhao2021efficientface,wang2020ran,li2018gacnn} and vision transformers (ViT) \cite{li2021mvt,dosovitskiy2020vit}. Besides, many hybrid models utilize CNNs to extract local features and apply ViT to model global dependencies. For example, Ma et al. \cite{ma2021vtff} used a two-branch CNN to extract features from the original image and its LBP, then fed the fused features to a ViT to make the final inference. Xue et al. \cite{xue2021transfer} devised two dropping strategies to randomly remove self-attention module or erase the attention maps of ViT, thereby forcing the network to extract comprehensive local information. As for DFER, the spatial features of each frame are usually learned by CNNs \cite{simonyan2014vgg,he2016resnet,wang2022dpcnet}, and then the temporal dependencies are modeled by RNNs \cite{zhang2018RNN,lu2018rnn,wang2019rnn} or LSTM \cite{hochreiter1997lstm}. In recent years, some researchers utilize transformer \cite{vaswani2017transformer} to model the long-range dependencies between the frames. Among them, Zhao et al. \cite{zhao2021former} introduced a Former-DFER, which learns both spatial and temporal features by transformers, and Ma et al. \cite{ma2022stt} proposed a spatio-temporal transformer (STT) to jointly learn the spatial and temporal features. 

The above methods all take one-hot \cite{wang2020scn}, or soft labels obtained through teacher networks or multi-person annotations \cite{zhao2021efficientface} as the supervision to train their models, which lack semantic information and are less interpretable. In addition, different people express the same expression in ever-changing ways, and even the same person can have many different ways to express the same expression. So, it is unreasonable to give equal supervision for all the samples of the same expression (i.e., one-hot label). Therefore, we propose METD to learn more fine-grained representations of expressions by introducing learnable expression text descriptors as the supervision.

\subsection{Vision-Language Models}
\label{sec:2.2}
Recently, vision-language models \cite{radford2021clip} have greatly improved the zero-shot performance by conducting contrastive learning on image-text pairs. These image-text pairs collected from the Internet are almost infinite, and collecting them is very efficient and costless. The most recent vision-language learning frameworks CLIP \cite{radford2021clip} and ALIGN \cite{jia2021align} benefit from 400 million and 1.8 billion image-text pairs collecting from the Internet. These large-scale VLP models are then applied to various downstream CV tasks and achieve SOTA performance, such as object detection \cite{xie2021zsdet}, semantic segmentation \cite{xu2021zsseg}, and video understanding \cite{xu2021videoclip}. 

Unlike the above-mentioned tasks that focus on understanding macroscopic objects, expression-related text representations are often not limited to facial expressions. Visualization results in \cite{goh2021clipvisual} have shown that most expression-related semantics in CLIP are complex and compound concepts (e.g., ``angry'' is related to ``evil'', ``serious'', and ``mental health disorder''). In summary, the expression-related concepts in the text are abstract and compound, so it is tedious and impractical to design 
handcrafted text prompts for facial expression. Therefore, we devise METD, avoiding hand-engineering for expression text prompts.
\begin{figure*}[t]

  \centering
  \includegraphics[width=14cm]{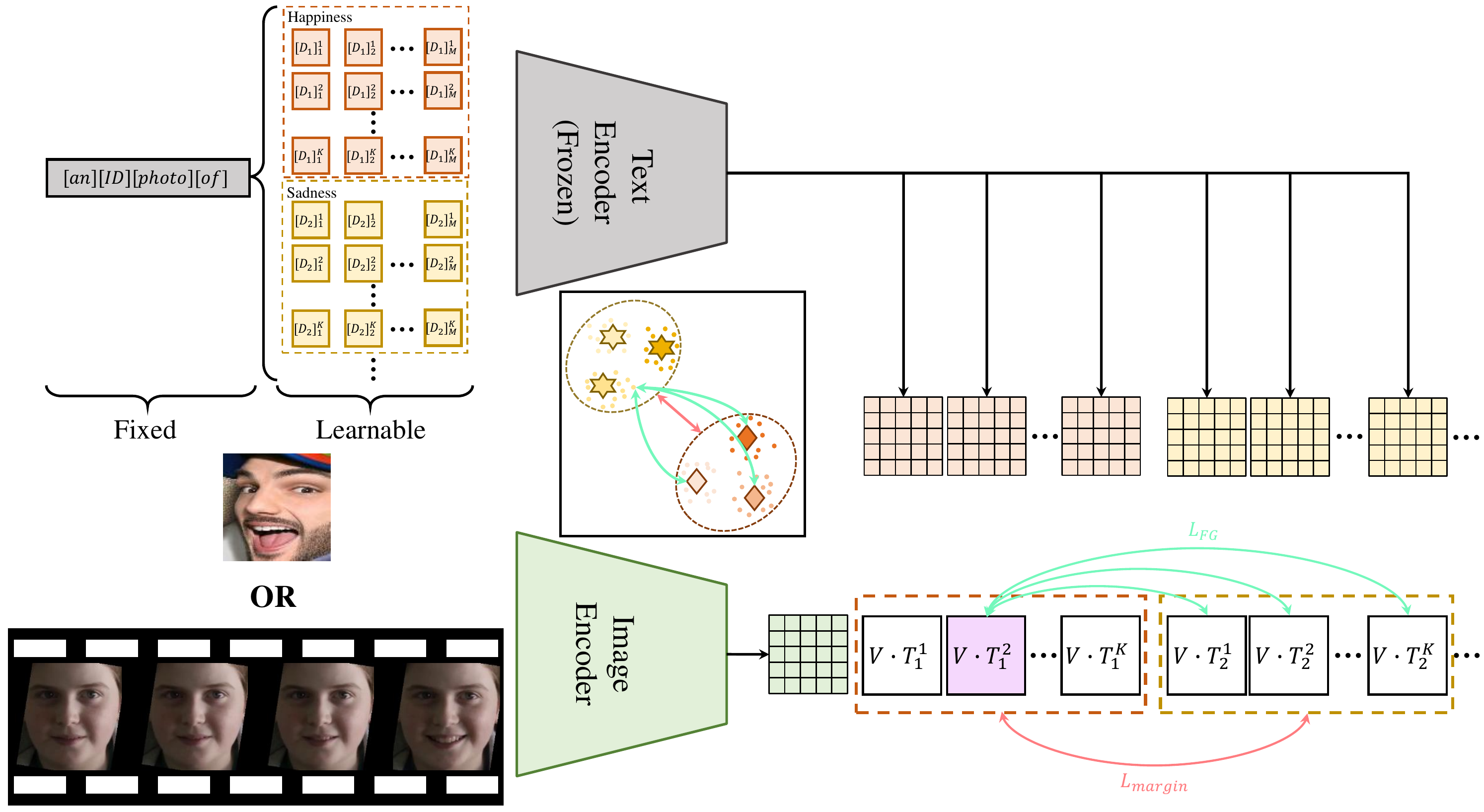}
  \caption{The pipeline of CLIPER. $V$ stands for the image embedding generated by the image encoder $\mathcal{F}_I$. $T_i^k$ denotes the text embeddings generated from the $k$-th text descriptor of the $i$-th expression by the text encoder $\mathcal{F}_T$.}
 \label{Fig:pipline}
\end{figure*}
\subsection{Prompt Learning}\label{sec:2.3}
Knowledge probing as the predecessor of prompt learning is first proposed in \cite{petroni2019prompt}. Then, researchers begin to use prompt to extract knowledge from large-scale language pre-trained models, such as BERT \cite{devlin2018bert} and GPT \cite{radford2019gpt}, to deal with natural language processing (NLP) tasks \cite{shin2020autoprompt,li2021prefix}. 

With the development of contrastive vision-language pre-training models \cite{li2022blip,radford2021clip} that maps the text and image embedding into a shared space, researchers usually utilize a prompt (e.g., ``a photo of a $\textbf{\{class\}}$") along with the name of class to generate a text descriptor for each class. After that, the text embeddings generated from these descriptors are used to make zero-shot inferences on downstream tasks. However, the fixed prompt cannot be optimal for all the downstream tasks. For example, for the flower classification task, ``a photo of a {flower class}, a type of flower'' can be a better prompt than the standard prompt. 

To address this issue, Zhou et al. \cite{zhou2022coop} introduced a learnable continuous prompt method named context optimization (CoOp), which can adaptively adjust according to different datasets. Furthermore, Zhou et el. \cite{zhou2022cocoop} proposed conditional context optimization (CoCoOp) based on CoOp, which can generate a prompt conditioned on each image and make better inferences on unseen classes. These two methods both focus on learning a continuous context for generating a better text descriptor along with the name of the class, such as ``cat'', ``car'', and ``flower''. However, as we mentioned in Section~\ref{sec:2.2}, expression-related text is compound and complex, thus cannot be used as a good prior for generating text descriptors of facial expressions. Therefore, we propose multiple expression text descriptors (METD), which generates a set of learnable text descriptors for each expression without using names of the facial expressions.


\section{Method}
\subsection{Preliminaries}
Since we are the first to apply contrastive vision-language pre-training models to the in-the-wild FER tasks, we will briefly introduce the backbone network CLIP \cite{radford2021clip} used in this paper. 

\paragraph{Model and Training} CLIP is a contrastive multi-modal pre-training framework that maps the text and image features into a shared space by deploying contrastive learning with 400 million text-image pairs collected from the Internet. As shown in Figure~\ref{Fig1}(b), CLIP consists of two independent encoders, i.e., $\mathcal{F}_I$ and $\mathcal{F}_T$, used to map the image and the text input into a shared embedding space. During the training process, CLIP maximizes the cosine similarity for features of the matched image-text pairs while minimizing that of all other negative pairs. 

\paragraph{Inference} Common classification usually map the learned features into a $N$-dimensional vectors $y^{pred}$ for inference. $N$ denotes the number of classes. The prediction probability of input image $x$ on the $i$-th class can be formulated as,

\begin{equation}\label{eq:1}
P(y=i|x)=\frac{e^{y_{i}^{pred}}}{ {\textstyle \sum_{j=1}^{N}e^{y_{j}^{pred} } } },
\end{equation}
\noindent where $y_{i}^{pred}$ is the output logits of the $i$-th class.

Since CLIP maps both the text and the image embedding into an aligned space, the cosine similarity of text and image embedding in the aligned space can be taken as the basis for inference. Specifically, CLIP uses the standard template (i.e., ``a photo of $\textbf{\{class\}}$'') to generate a text descriptor for each class. The prediction probability of a input image $x$ on the $i$-th class in CLIP can be calculated by,

\begin{equation}\label{eq:2}
P(y=i|x)=\frac{e^{\langle V, T_{i} \rangle/\tau }}{ {\textstyle \sum_{j=1}^{N}e^{\langle V, T_{j} \rangle/\tau }}},
\end{equation}
\noindent where
\begin{equation}\label{eq:3}
V=\mathcal{F}_I(x),
\end{equation}
\noindent and
\begin{equation}\label{eq:4}
T_i=\mathcal{F}_T(text_i).
\end{equation}

\noindent Where $\mathcal{F}_I$ and $\mathcal{F}_T$ are the image and the text encoder, respectively. $text_i$ denotes text descriptor of the $i$-th class. $\tau$ represents a temperature parameter of CLIP and $\langle\cdot, \cdot\rangle$ stands for the cosine similarity between two vectors. $N$ denotes the number of classes.

\subsection{CLIPER}
\subsubsection{Overview}
CLIP achieves impressive zero-shot inference performance on many object classification tasks. Each category in these classification tasks usually has a clear and uniform textual definition (e.g., ``dog'' and ``airplane''). However, the standard prompt is not the optimal text descriptor for the FER task, whose textual definitions of the categories, such as ``happy'' and ``sad'', are abstract and compound. To tackle the above issue, We propose a unified FER framework called CLIPER, which can automatically learn a set of expression text descriptors for each facial expression without adding any expression-related textual prior. 

As shown in Figure~\ref{Fig:pipline}, we set $K$ learnable expression text descriptors for each expression. Each learnable text descriptor consists of a fixed part, i.e., ``an ID photo of'' and $M$ learnable text tokens used to describe the facial expressions. These $N\times K$ text descriptors are fed into the text encoder to get $N$ groups of text embeddings corresponding to $N$ expressions. As for the visual parts, the image is fed to the image encoder to obtain the image embedding $V$. Specially, the visual embedding of a video sequence is obtained by aggregating the image embedding of each frame through mean pooling, which also proves that the gain of CLIPER on DFER task does not come from the design of the temporal module. Finally, the mean cosine similarities of each expression group are used to make the inference.
\subsubsection{Two-Stage Training Paradigm}
The training process of our model can be divided into two stages. At the first stage, CLIPER learns the proposed multiple expression text descriptors (METD) from the aligned vision-text embedding space. Specifically, the text descriptor of the $k$-th subclass of the $i$-th superclass is in the following form,
\begin{equation}\label{eq:5}
text_i^k = [\rm{Context}][D_i]_1^k[D_i]_2^k\cdot\cdot\cdot[D_i]_M^k.
\end{equation}
Where $k\in \left\{1,2,\cdot\cdot\cdot ,K \right \}$. $[D_i]_m^k$ stands for the $m$-th learnable text token belongs to the $k$-th subclass of the $i$-th superclass, as shown in Figure~\ref{Fig:pipline}. $M$ and $K$ are the length of the learnable text token and number of the subclasses in each superclass. It should be noted that METD focuses on learning text tokens that describe expressions, while other prompt learning methods, such as CoOp and CoCoOp are designed to learn class irrelevant context part, i.e., $[\rm{Context}]$ in Eq~\ref{eq:5}, which is fixed in our method.

To keep the aligned embedding space, the parameters of both the text and the image encoders need to be frozen at this stage. For the loss function, we devise a novel fine-grained cross-entropy loss to maximize the cosine similarity between the image embedding and the closest text embedding of the target superclass, and minimize the similarities between the image embedding and all the text embeddings from other non-target superclasses. For a visual input of the $t$-th class, the fine-grained cross-entropy loss can be formulated as, 
\begin{equation}\label{eq:6}\small
\mathscr{L}_{FG}=-\alpha^{+}_t\cdot log\frac{e^{\langle V, T^{+}_t \rangle/\tau }}{e^{\langle V, T^{+}_t \rangle/\tau }+ \sum^{N}_{i=1\, i\ne  t}\sum_{k=1}^{K}e^{\langle V, T^{k}_i \rangle/\tau }},
\end{equation}
with 
\begin{equation}\label{eq:7}
\alpha^{+}_t = \frac{e^{\frac{1}{n^{+}_t}}}{ {\textstyle \sum_{k=1\, n^{k}_t\ne0}^{K}e^{\frac{1}{n^{k}_t} }} } \frac{{\textstyle \sum_{k=1}^{K}n^{k}_t}}{n^{+}_t}.
\end{equation}

\noindent Where $T^{+}_t$ and $\alpha^{+}_t$ stand for the closest text embedding from the $K$ subclasses of the $t$-th superclass and the modulating factor corresponding to the closest subclass, respectively. The modulating factor is used to force the model learn more on the subclasses with less samples, thus preventing the model from collapsing to learn only one subclass for each expression superclass. $n^{k}_t$ and $n^{+}_t$ denote the number of the samples belongs to the $k$-th and the closest subclass of the $t$-th superclass, respectively. $T^{k}_i$ represents the text embedding from the $k$-th subclass of the $i$-th superclass, which can be calculated by,
\begin{equation}\label{eq:8}
T^{k}_i=\mathcal{F}_T(text_i^k),
\end{equation}
\noindent$\tau$ is a temperature
parameter of CLIP, and $V$ denotes the image embedding. 
As shown in Figure~\ref{Fig:pipline}, we hope CLIPER can keep a margin between different superclasses so that the boundaries between categories can be more clear. Therefore, we introduce a margin loss, which can be defined by,

\begin{equation}\label{eq:9}
\mathscr{L}_{margin}=-log\frac{e^{\langle V, T^{-}_t \rangle/\tau }}{e^{\langle V, T^{-}_t \rangle/\tau }+ \sum^{N}_{i=1\, i\ne  t}e^{\langle V, T_i^{+} \rangle/\tau }}.
\end{equation}

\noindent Where $T^{-}_t$ is the most dissimilar text embedding among the $K$ subclasses of the target superclass. This loss keeps the most dissimilar subclass in the target superclass still closer to the image embedding than the most similar subclass in other non-target superclasses. The overall loss function of CLIPER is defined as follows,

\begin{equation}\label{eq:10}
\begin{aligned}
\mathscr{L}_{all} =\mathscr{L}_{FG}+\mathscr{L}_{margin}.
\end{aligned}
\end{equation}

At the second stage, the learned $N\times K$ text descriptors are fixed. We only fine-tune the image encoder to extract more discriminative facial expression-related features. During the training process, the same loss function in Eq~\ref{eq:10} is used to guide the image encoder to learn more fine-grained features according to the METD learned in the first stage. Since METD learns multiple text descriptors for each expression, we use the mean value of the cosine similarity between all subclasses and the image embeddings when making the inference. Therefore, the prediction logits of the $i$-th expression can be formulated as,

\begin{equation}\label{eq:11}
\begin{aligned}
y^{pred}_i = \frac{\sum_{k=1}^{K}\langle V, T_i^{k} \rangle}{K}.
\end{aligned}
\end{equation}
\section{Experiments}
\subsection{Datasets}
\paragraph{SFER Datasets:} We choose two popular SFER benchmarks to evaluate the effectiveness of our method. \textbf{RAF-DB} \cite{li2017rafdb} is an in-the-wild facial expression dataset that contains 29,672 real-world facial expression images. The images are collected by image search API and annotated by 40 annotators. Consistent with most of the previous work \cite{wang2020scn,zhao2021efficientface}, only 15,339 images of seven basic expressions, i.e., surprise, fear, disgust, happy, sad, anger, and neutral, are used for training and testing. Among them, 12,271 images are used for training, and 3,068 images are for
testing. \textbf{AffectNet} \cite{mollahosseini2017affectnet} is the largest SFER dataset so far. It contains more than one million facial images collected from the Internet. While only about 450,000 images have been annotated manually with 11 expression categories. Following the setting in the previous works, we use 283,901 facial images of seven prototypical expressions, i.e., surprise, fear, disgust, happy, sad, anger, contempt, and neutral, as the training set and 4,000 images as the test set. It is worth noting that the training set is imbalanced but the test is balanced (i.e., 500 test images for each expression). 

\paragraph{DFER Datasets:} We conduct experiments on three popular dynamic facial expression recognition datasets to further evaluate our method. \textbf{DFEW} \cite{jiang2020dfew} consists of over 16,000 video clips collected from thousands of movies. Each video is individually annotated by ten annotators under professional guidance and assigned to one of the seven basic expressions (i.e., happiness, sadness, neutral, anger, surprise, disgust, and fear). Consistent with previous works, 12,059 video clips are used for conducting experiments. All the samples have been split into five same-size parts (fd1$\sim$ fd5) without overlap. The division of 5 folds is originally provided when obtaining the dataset. We choose 5-fold cross-validation, which takes one part of the samples for testing and the remaining for training, as the evaluation protocol. \textbf{AFEW} \cite{dhall2012afew} served as an evaluation platform for the annual EmotiW from 2013 to 2019 that contains 1,809 video clips collected from movies. All the clips are split into three non-overlapped subsets (i.e., training (773 video clips), validation (383 video clips), and testing (653 video clips)). Since the training set of AFEW has a very limited scale, we pre-train our model on the fd1 of the DFEW and then fine-tune the model on the training set of AFEW as done in \cite{zhao2021former,ma2022stt,li2022ial}. Besides, since the test split is not publicly available, we report results on the validation set as the previous work did. \textbf{FERV39K} \cite{wang2022ferv39k} is currently the largest in-the-wild DFER dataset and contains 38,935 video sequences collected from 4 scenarios, which can be further divided into 22 fine-grained scenarios, such as crime, daily life, speech, and war. Each clip is annotated by 30 individual annotators and assigned to one of the seven basic expressions as DFEW. All the samples are divided into a training set (80\%/31,088 clips) and a test set (20\%/7,847 clips) without overlapping.

\paragraph{Validation Metrics:} Consistent with most previous methods \cite{wang2020scn,zhao2021former}, for SFER, we take the weighted average recall (WAR, i.e., accuracy) as the validation metric. As for DFER, both WAR and the unweighted average recall (UAR, i.e., the accuracy per class divided by the number of classes without considering the number of instances per class) are utilized for evaluating the performance of methods.

\subsection{Implementation Details}
All the images and video frames in our experiments are resized to $224\times 224$. The random resized crop, random erasing, horizontal flipping, and color jittering are used to avoid overfitting. For all the SFER datasets, ViT-B/16 of CLIP \cite{radford2021clip} is used as the backbone of CLIPER. The AdamW \cite{loshchilov2018adamw} optimizer is used to minimize the loss function with a mini-batch size of 128. The initial learning rate is set at $1\times 10^{-2}$ for RAF-DB and $5\times 10^{-5}$ for AffectNet at the first stage to optimize the METD for two epochs with no weight decay and set at $5\times 10^{-6}$ to fine-tune the image encoder at the second stage for 50 epochs. For all the DFER datasets, since extracting features from video sequences are more expensive, we use the ViT-B/32 of CLIP \cite{radford2021clip} as the backbone of CLIPER. For sampling frames, we sampled 16 frames from each video clip following the sampling strategy in \cite{li2022ial,ma2022stt,zhao2021former}. The AdamW optimizer is used with a mini-batch size of 48. The initial learning rate is set at $1\times 10^{-2}$ at the first stage to optimize the METD for one epoch with no weight decay and set at $5\times 10^{-6}$ to fine-tune the image encoder at the second stage for 36 epochs. 

At the second stage, the learning rate is cosine decayed \cite{loshchilov2016sgdr}. The weight decay is set at 0.1 for both SFER and DFER datasets at the second stage. Specially, since pre-training weights of CLIP are half-precision, the eps of all optimizers is set to $1\times 10^{-4}$ for stable training. The temperature parameter $\tau$ of CLIP is set at 0.01. By default, the number of text descriptors for each expression $K$ is set at 5, and the length of the learnable text tokens is set at 4. Specially, we take the mean value of the cosine similarities of all subclasses for inference, as described in Eq~\ref{eq:11}. All the experiments are conducted on one NVIDIA V100 GPU with PyTorch toolbox.

\begin{table}
\begin{center}

\begin{tabular}{c|cc|c}
\toprule
\multirow{2}{*}{Method}&\multicolumn{2}{c|}{FERV39K (\%)}&\multicolumn{1}{c}{RAF-DB (\%)}\cr
    \cmidrule(lr){2-4}
    & UAR & WAR& WAR\cr
\midrule
CLIP (zero-shot)  &21.20 &16.20&40.16\cr
CoOp \cite{zhou2022coop}&32.40&43.57 &84.65\cr
CoCoOp \cite{zhou2022cocoop}&34.32 &45.53&85.89\cr
\midrule
CLIPER (Ours)  &\textbf{41.23} &\textbf{51.34}&\textbf{91.61} \cr

\bottomrule
\end{tabular}\caption{Comparison of several continuous prompt learning methods (i.e., the proposed CLIPER, CoOp, and CoCoOp). CLIP: ``an ID photo of a $\{\textbf{expression}\}$ face'', CoOp and CoCoOp: ``$X X X X$ $\{\textbf{expression}\}$'', and CLIPER: ``an ID photo of $X X X X$''. $X$ stands for a learnable text token.
}\label{tab:pl}
\end{center}
\end{table}
\begin{table}
\begin{center}

\begin{tabular}{c|cc|c}
\toprule
\multirow{2}{*}{Strategy}&\multicolumn{2}{c|}{FERV39K (\%)}&\multicolumn{1}{c}{RAF-DB (\%)}\cr
    \cmidrule(lr){2-4}
    & UAR & WAR& WAR\cr
\midrule
CLIP (zero-shot)  &21.20 &16.20&40.16\cr
Linear Probe &33.06&44.57 &84.29\cr
Fully Fine-tuning &38.68 &50.13&90.38\cr
\midrule
Two-stages (Ours)  &\textbf{39.32} &\textbf{50.52}&\textbf{90.91} \cr

\bottomrule
\end{tabular}\caption{Comparison of different fine-tuning strategies. ``Linear Probe'': adding a linear classifier to the image encoder and optimizing the parameters of the linear layer with image encoder fixed. ``Fully fine-tuning'': fine-tuning both the linear layer and the image encoder. Both strategies do not introduce the text encoder.
}\label{tab:ft}
\end{center}
\end{table}
\subsection{Ablation Studies}
We conduct ablation studies on both SFER and DFER datasets to demonstrate the effectiveness of CLIPER and prove our motivation. Specifically, we select RAF-DB \cite{li2017rafdb}, and FERV39K \cite{wang2022ferv39k} to support all the ablation studies.

\paragraph{Comparison with Other Prompt Learning Methods} The most relevant works to CLIPER are CoOp \cite{zhou2022coop} and CoCoOp \cite{zhou2022cocoop}, both of which address the problem of prompt engineering when transferring large-scale contrastive vision-language  pre-training models to downstream tasks. Among them, CoOp tries to learn continuous context for each downstream task to replace the unified standard prompts in CLIP (i.e., ``a photo of a $\{\textbf{class}\}$''), so that the continuous prompts can adapt to different downstream tasks. While CoCoOp aims to adaptively generate a set of specialized continuous contexts for each sample so that CLIP can better generalize to the unseen categories. Although CoOp and CoCoOp learn a continuous context for downstream tasks, they still need the name of the class (e.g., ``cat'' and ``dog'') to generate complete text descriptors to make inferences. However, as declared in Section~\ref{sec:2.2} and Section~\ref{sec:2.3}, expression-related texts, such as ``happy'' and ``angry'', are compound and complex, thus cannot be taken as a good prior to generate the text descriptor for FER tasks. Therefore, we proposed METD to learn continuous text descriptors to describe the specific expression class rather than directly using the name of the class. To demonstrate our point, we compare CLIPER with the above two methods in Table~\ref{tab:pl}, and also give the zero-shot inference performance of CLIP as a baseline model for comparison. For a relatively fair comparison, the number of text tokens $M$ is set to 4 for CLIPER, CoOp, and CoCoOp. From Table~\ref{tab:pl}, we can see that our CLIPER significantly outperforms the above prompt learning methods, which further indicates that learning the text descriptors for each expression is much more important than do prompt engineering on the context as did in CoOp and CoCoOp.

\begin{table}[t]
\begin{center}

\begin{tabular}{c|c|cc|c}
\toprule
\multicolumn{2}{c|}{Setting}&\multicolumn{2}{c|}{FERV39k (\%)}&\multicolumn{1}{c}{RAF-DB (\%)}\cr
    \cmidrule(lr){1-5}$K$&$M$
    & UAR & WAR& WAR\cr
\midrule
1 &4 &39.32& 50.52                           &90.91\cr
5 &4 &\textbf{41.23}& 51.34                           &\textbf{91.61}  \cr
10 &4 &39.49&  51.00                         &91.36\cr
1 &8 &40.37&50.86&90.65\cr
5 &8 &40.62& 51.13 &91.20 \cr
10 &8 &40.78& \textbf{51.50}&91.17 \cr
\bottomrule
\end{tabular}\caption{Evaluation of different number of the subclasses $K$ and the length of the learnable text tokens $M$.
}
\label{tab:hp}
\end{center}
\end{table}
\paragraph{Comparisons of Different Fine-Tuning Strategies} We propose a two-stage training paradigm to transfer CLIP to the FER task. This paradigm mainly bases on two insights: 1) The potential of CLIP pre-training weights can only be fully exploited by being guided by text descriptors rather than directly fine-tuning the image encoder with a fully connected (FC) layer; 2) For the FER task, the image encoder needs to be fine-tuned, because original CLIP focuses on modeling object-related semantics (i.e., ``face'') rather than facial texture details related to expressions (i.e., what happens on ``face''). Therefore, CLIPER learns the multiple expression text descriptors while keeping the text-image aligned embedding space at the first stage and then fine-tuning the image encoder to learn more facial texture details related to expressions. To support the above insights, we try different fine-tuning strategies in Table~\ref{tab:ft}. For a fair comparison, we only use the results of a single expression text descriptor (i.e., $K=1$), so the results will not benefit from METD. Since the default learning rate may lead to training failure in some strategies (due to the mixed precision weight), we slightly adjust the learning rate used in some strategies so that the comparison is relatively fair. As shown in Table~\ref{tab:ft}, our two-stage fine-tune paradigm outperforms the ``fully fine-tuning'', which demonstrates our first insight, i.e., introducing text to guild the fine-tuning process of image encoder is helpful. Besides, the ``fully fine-tuning'' significantly exceeds the results using ``linear probe'', which corresponds to the second insight that the origin image encoder does not learn enough facial texture details related to expressions.

\paragraph{Evaluation of Hyper-Parameters in METD} The length of the learnable text tokens $M$ and the number of the subclasses in each expression superclass $K$ are two main hyper-parameters of the proposed METD. As shown in Table~\ref{tab:hp}, the results of using METD outperform using a single text descriptor (i.e., $K=1$) for each class. In addition, a larger length seems more suitable for the DFER datasets with more data. While for the SFER dataset with less data, a smaller length can prevent overfitting.

\begin{table}[t]
\begin{center}

\begin{tabular}{c|c|c}
\toprule
Method&RAF-DB (\%)&AffectNet-7/8$^\dagger$ (\%)\cr
    
\midrule
DLP-CNN \cite{li2017rafdb}  &80.89 &54.47 \cr
gACNN \cite{li2018gacnn} &85.07&58.78\cr
IPA2LT \cite{zeng2018ipa2lt}&86.77&55.71\cr
RAN \cite{wang2020ran}&86.90&52.97$^\dagger$\cr
SCN \cite{wang2020scn}&87.03&60.23$^\dagger$\cr
KTN \cite{li2021ktn}&88.07&63.97\cr
EfficientFace \cite{zhao2021efficientface}&88.36&63.70/59.89$^\dagger$\cr
MVT \cite{li2021mvt}&88.62&64.57/61.40$^\dagger$\cr
VTFF \cite{ma2021vtff}&88.14&64.80/61.85$^\dagger$\cr
TransFER \cite{xue2021transfer}&90.91&66.23\cr
 Face2Exp \cite{zeng2022face2exp}&88.54&64.23\cr
\midrule
CLIPER (Ours) &\textbf{91.61}&\textbf{66.29}/\textbf{61.98}$^\dagger$\cr

\bottomrule
\end{tabular}\caption{Comparison with state-of-the-art methods on RAF-DB and AffectNet datasets. The bold denotes the best. For AffectNet, training and testing are on 7 classes without ``contempt'' by default. $^\dagger$ denotes training and testing on 8 classes with ``contempt''.
}
\label{tab:sfer}
\end{center}
\end{table}

\begin{table*}[t]
\begin{center}

\begin{tabular}{c|cc|cc|cc}
\toprule
\multirow{2}{*}{Method}&\multicolumn{2}{c|}{FERV39k (\%)}&\multicolumn{2}{c|}{DFEW (\%)}&\multicolumn{2}{c}{AFEW (\%)}\cr
    \cmidrule(lr){2-7}& UAR & WAR& UAR & WAR& UAR & WAR\cr
    
\midrule
I3D-RGB \cite{carreira2017I3D-RGB} &30.17 &38.78&46.52&58.27&41.86&45.41 \cr
3D ResNet18 \cite{hara20183dres}  &26.67 &37.57&46.52&58.27&42.14&45.67 \cr
EC-STFL \cite{jiang2020dfew}&-- &--&45.35&56.51&--&53.26\cr
Former-DFER \cite{zhao2021former} &37.20 &46.85&53.69&65.70&47.42&50.92 \cr
STT \cite{ma2022stt}&37.76    &48.11&54.58&66.65&49.11&54.23\cr
NR-DFERNet \cite{li2022nrdfer}&33.99&45.97&54.21&68.19&48.37&53.54\cr
DPCNet \cite{wang2022dpcnet} &--&--&57.11&66.32&47.86&51.67\cr
IAL \cite{li2022ial} &35.82&48.54&55.71&69.24&50.84&54.59\cr
\midrule

CLIPER (Ours)&\textbf{41.23}&\textbf{51.34}&\textbf{57.56}&\textbf{70.84}&\textbf{52.00}&\textbf{56.43}\cr
\bottomrule
\end{tabular}
\caption{Comparison with state-of-the-art methods on AFEW, DFEW, and FERV39K datasets. The best results are highlighted in bold. 
}
\label{tab:dfer}
\end{center}
\end{table*}
\subsection{Comparison with State-of-the-Arts}
In this section, we compare the proposed CLIPER with current state-of-the-art methods on both DFER (i.e., AFEW \cite{dhall2012afew}, FERV39K \cite{wang2022ferv39k}, and DFEW \cite{jiang2020dfew}) and SFER (i.e., RAF-DB \cite{li2017rafdb} and AffectNet \cite{mollahosseini2017affectnet}) benchmarks to demonstrate the superiority of our methods.

Table~\ref{tab:sfer} compares CLIPER to several SOTA SFER methods on two popular SFER datasets. It can be seen that our method outperforms other SOTAs on both datasets. Specially, since the training set of AffectNet is imbalanced, we use the oversampling strategy to force the network to learn more on the classes with less samples as did in \cite{li2021mvt,zhao2021efficientface,ma2021vtff,wang2020scn}.

Table~\ref{tab:dfer} shows the results of UAR/WAR on three mainstream DFER benchmarks. We can see that CLIPER significantly exceeds the performance of other DFER SOTAs on three datasets. It should be noted that CLIPER aggregate the features of the frames through the most naive way (i.e., temporal mean pooling), which means that CLIPER hardly benefits from the design of temporal modules. This makes CLIPER still have great potential and can serve as a strong baseline model for extracting facial features.

\begin{table}[t]
\begin{center}

\begin{tabular}{l|c}
\toprule
EXP&Index and Closest ``Word''\cr
    
\midrule
Surprise &(2,3) zombies; (3,3) possibly; (4,4) nugget\cr
Fear &(1,3) ``!''; (3,4) ned; (5,3) indictment\cr
Disgust &(2,1) sanctions; (2,2) fern; (4,1) boredom\cr
Happy  &(2,2) benefits; (3,1) kindness; (4,2) hopeful \cr
Sad &(1,1) defeats; (2,3) saddest; (3,3) smear\cr
Anger &(1,3) berser; (3,1) roaring; (3,2) slam\cr
Neutral &(2,2) abstraction; (3,2) relaxation; (4,1) loose\cr

\bottomrule
\end{tabular}\caption{The nearest words for the multiple expression text descriptors learned by CLIPER. Each of the seven basic expressions correspond to $K\times M=20$ nearest words, when $K=5$ and $M=4$. Due to the space limitation, we only show some nearest words and give their corresponding index $(k,m)$, which means the $m$-th learned text token of the $k$-th subclass of the corresponding superclass (i.e., $[D]_m^k$). ``EXP'' denotes expression.
}
\label{tab:words}
\end{center}
\end{table}

\subsection{Visualization}
CLIPER not only achieves good performance on both DFER and SFER datasets, but also has good interpretability compared to other FER methods. Here we present some visualization results on RAF-DB dataset to demonstrate the interpretability of CLIPER.

\paragraph{What Text Token Does CLIPER Learn?} CoOp \cite{zhou2022coop} give some visualizations about what the learned continuous text token is by searching within the vocabulary for words that are closest to the learned vectors based on the Euclidean distance. 
We show some results of the closest words in the dictionary learned by METD in Table~\ref{tab:words}. From these results, we can see that our METD learns not only some texts directly related to expressions (e.g., ``saddest'' for sad and ``kindness'' for happy), but also some indirectly related semantics (e.g., ``roaring'' for angry and ``zombies'' for surprise), which demonstrates the interpretability of our METD. We provide the full word list in supplementary materials.
\begin{figure}[t]
  \centering
  \includegraphics[width=6.5cm,height=6.5cm]{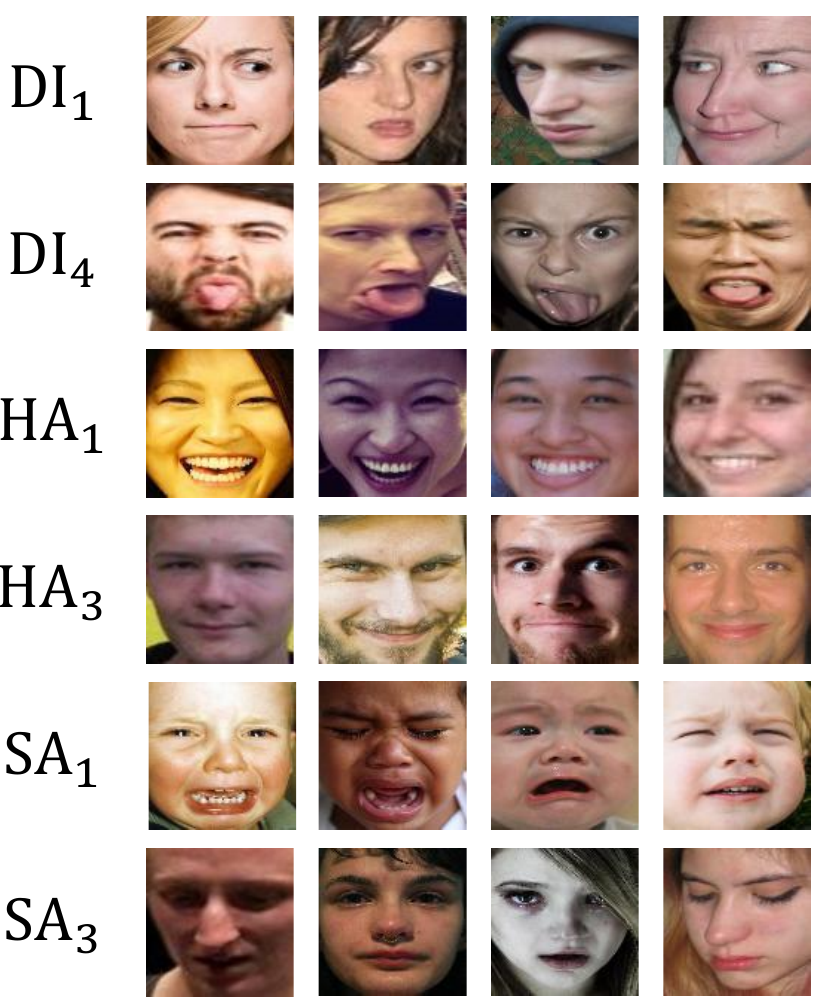}
  \caption{Visualizations of some samples in different subclasses of several expressions. DI, HA, and SA represent disgust, happy, and sad, respectively. The subscript indicates the index of the subclass.}
 \label{Fig:fg}
\end{figure}
\paragraph{Does METD Learn More Fine-Grained Expression Semantics?} The original intention of designing METD is to learn diverse forms of expressions. We show some facial images with their corresponding subclass index from the test set of RAF-DB dataset in Figure~\ref{Fig:fg}. The results show that METD has learned some fine-grained expression representations. For ``disgust'', METD learns different forms to express these two expressions. For ``happy'', METD divide the subclass $\rm HA_1$ and $\rm  HA_3$ according to the intensity of ``happy'', i.e., smile or laugh. While for ``sad'', METD seems to divide the subclass based on age.

\section{Discussion}
It should be noticed that CLIP can obtain very high zero-shot performance on many general classification datasets, such as CIFAR10 \cite{krizhevsky2009cifar} and ImageNet \cite{russakovsky2015imagenet}. These datasets have one thing in common, that is, the categories in the dataset have a clear and uniform definition and can usually be described by a noun (e.g., car and dog). For some datasets that consists of abstract classes, CLIP often do not performs well, such as MNIST \cite{lecun1998mnist} and FER datasets. The main reason is that these categories often correspond to multiple concepts. For MNIST dataset, ``one'' can mean ``one pen'' or ``one car''. For the FER datasets, ``happy'' is highly related to ``birthday'', ``champion'', and etc concepts \cite{goh2021clipvisual}. The text descriptors generated from these compound concept (e.g.,``a photo of happy'') are harmful and leading to confusion in the FER task, which also explains why CLIP achieve poor zero-shot performance on FER datasets. Therefore, CLIPER adaptively learns more fine-grained text descriptors through the proposed two-stage training paradigm, avoiding the direct use of class names of the FER datasets.

\section{Conclusion}
We develop CLIPER based on CLIP, a unified FER framework. It can be applied to both in-the-wild SFER and DFER tasks. By introducing multiple expression text descriptors (METD) to guide the fine-tuning process of the image encoder, CLIPER is more interpretable than previous FER approaches that take one-hot or soft labels as the supervision. Besides, METD can help CLIPER find more fine-grained subclasses of each expression. Extensive experiments demonstrate the effectiveness of CLIPER, which can serve as a strong baseline for FER tasks. The visualization results further certify the interpretability of our method.
\bibliographystyle{IEEEtran}
\bibliography{conference_101719}
\end{document}